\documentclass[10pt,twocolumn,letterpaper]{article}

\usepackage{cvpr}
\usepackage{times}
\usepackage{epsfig}
\usepackage{graphicx}
\usepackage{amsmath}
\usepackage{amssymb}
\usepackage{authblk}
\usepackage[breaklinks=true,bookmarks=false]{hyperref}

\cvprfinalcopy % *** Uncomment this line for the final submission

\def\cvprPaperID{****} % *** Enter the CVPR Paper ID here

\begin{document}

%%%%%%%%% TITLE
\title{Deep Keyframe Detection in Human Action Videos}

%\and
\author[1,2]{Xiang Yan}
\author[2]{Syed Zulqarnain Gilani}
\author[1]{Hanlin Qin}
\author[3]{Mingtao Feng}
\author[4]{Liang Zhang}
\author[2]{Ajmal Mian}

\affil[1]{School of Physics and Optoelectronic Engineering, Xidian University, China}

\affil[ ]{xyan@stu.xidian.edu.cn, hlqin@mail.xidian.edu.cn}

\affil[2]{School of Computer Science and Software Engineering, University of Western Australia}
\affil[ ]{{\{zulqarnain.gilani, ajmal.mian\}@uwa.edu.au}}

%\affil[ ]{\textit {\{zulqarnain.gilani,ajmal.mian\}@uwa.edu.au}}
\affil[3]{College of Electrical and Information Engineering, Hunan University, China}

\affil[ ]{mintfeng@hnu.edu.cn}

\affil[4]{School of Software, Xidian University, China}

\affil[ ]{liangzhang@xidian.edu.cn}

\maketitle
%\thispagestyle{empty}

%%%%%%%%% ABSTRACT
\begin{abstract}
   Detecting representative frames in videos based on human actions is quite challenging because of the  combined factors of  human pose in action and the background. This paper addresses this problem and formulates the key frame detection as one of finding the video frames that optimally maximally contribute to differentiating the underlying action category from all other categories. To this end, we introduce a deep two-stream ConvNet for key frame detection in videos that learns to directly predict the location of key frames. Our key idea is to automatically generate labeled data for the CNN learning using a supervised linear discriminant method. While the training data is generated taking many different human action videos into account, the trained CNN can predict the importance of frames from a single video. We specify a new ConvNet framework, consisting of a summarizer and discriminator. The summarizer is a two-stream ConvNet aimed at, first, capturing the appearance and motion features of video frames, and then encoding the obtained appearance and motion features for video representation. The discriminator is a fitting function aimed at distinguishing between the key frames and others in the video. We conduct experiments on a challenging human action dataset UCF101 and show that our method can detect key frames with high accuracy.
\end{abstract}

%%%%%%%%% BODY TEXT
\section{Introduction}

he ever increasing number of video cameras are generating unprecedenting amounts of big data that needs efficient algorithms for analysis. In particular, a large proportion of these videos depict events about people such as their activities and behaviors. To effectively interpret this data requires computer vision algorithms that have the ability to understand and recognize human actions. Its core task is to find out the compelling human action frames in videos which are able to well represent the distinct actions for their corresponding videos. Based on this insight, several papers have presented key frame based action recognition and video summarization approaches~\cite{zhang2016summary,gong2014diverse,mundur2006keyframe,zhu2016key,Kar_2017_CVPR,Mahasseni_2017_CVPR}. Therefore, detecting compelling key frames in human action videos has been crucial for video interpretation.
%------------------------------------------
  \begin{figure}
 \begin{center}
   %\includegraphics[width=0.8\linewidth]{egfigure.eps}
  \includegraphics[width=1.0\linewidth]{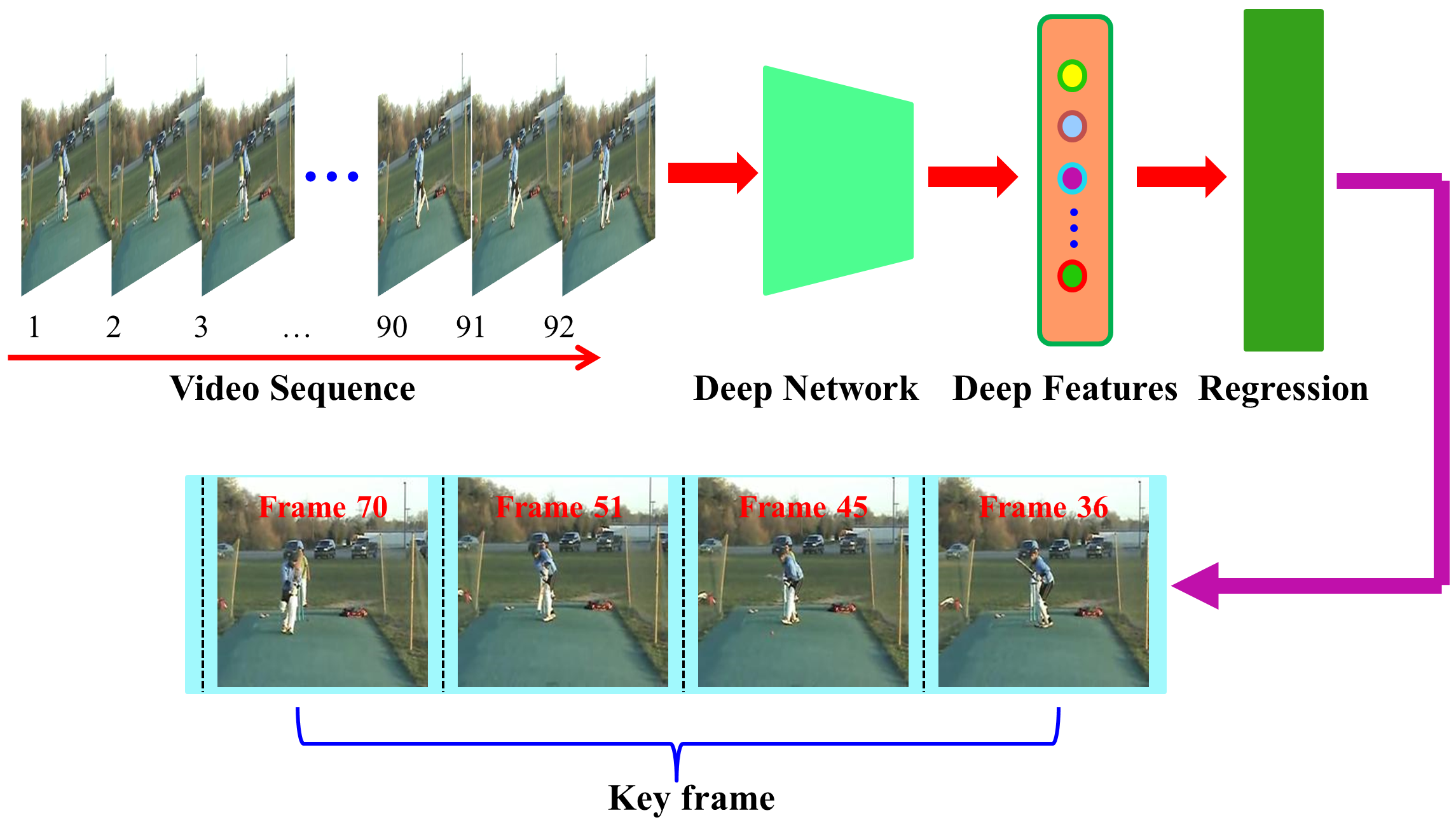}
  \end{center}
   \caption{Key frames detected by our deep keyframe detection network.}
  \label{fig:1}
    \end{figure}
%------------------------------------------

Early researchers have proposed various key frame detection methods by several strategies: segment-based~\cite{liu2003novel,lowe2004distinctive}
cluttering~\cite{sze2005new} based. Recently, inspired by the remarkable successes of deep convolutional neural networks (CNNs)  for image classification, object classification, and complex events detection~\cite{krizhevsky2012imagenet,simonyan2014very,tang2015improving,xiong2015recognize,karpathy2014large,yue2015beyond}, several key frame based works ~\cite{yang2015unsupervised,Mahasseni_2017_CVPR,Kar_2017_CVPR} have explored deep CNNs for key frame detection in human action videos.

However, most of these key frame based video understanding methods have not provided the ground truth key frames in human action videos. To the best of our knowledge, the only works that have mentioned the ground truth of key frames \cite{guan2013keypoint,ioannidis2016weighted}, report the key frames that are manually selected by two or three people with video processing background. Since this method is biased by human subjective factors, the ground truth of key frame selected by this method may not be consistent and may not be in line with how computer algorithms consider as key frames. Moreover, little attention has been devoted to the task of automatically generating ground truth frame for key frame detection in videos. This raises an interesting yet challenging question, how we can automate the process for key frame labeling in human action videos.

In this work, we introduce a automatic label generation algorithm that learns to discriminate each frame for key frame detection. Based on the frame-level video label, we devise a deep key frame detection in human action videos network that reasons directly on the entire video frames. Our key intuition is that the human action video can be well interpreted by only several compelling frames in a video. Based on this intuition, we draw support from the powerful feature representation ability of CNNs for images and the class separation ability and dimensionality reduction of linear discriminant analysis (LDA), we devise a novel automatically generating labels for training the deep key frame detection model. Our key frame detection model takes a long human action video as input, and output the key frames (compelling frames) of a action instance, as shown in Figure~\ref{fig:1}.

The core idea of this paper is to introduce a model that automatically output key frames from human action videos. Specifically, the contributions are threefold. (1) We introduce a novel deep two-stream ConvNets based key frame detection model, which combines the advantages of CNN model and LDA. (2) We also introduce a novel automatically generating label method to train the deep key frame detection model. (3) When applied to  UCF101~\cite{soomro2012ucf101}, the most popular action recognition dataset, our deep two-stream ConvNets model achieves excellent performance.
%-------------------------------------------------------------------------
\begin{figure*}
\begin{center}

\includegraphics[width=0.95\linewidth]{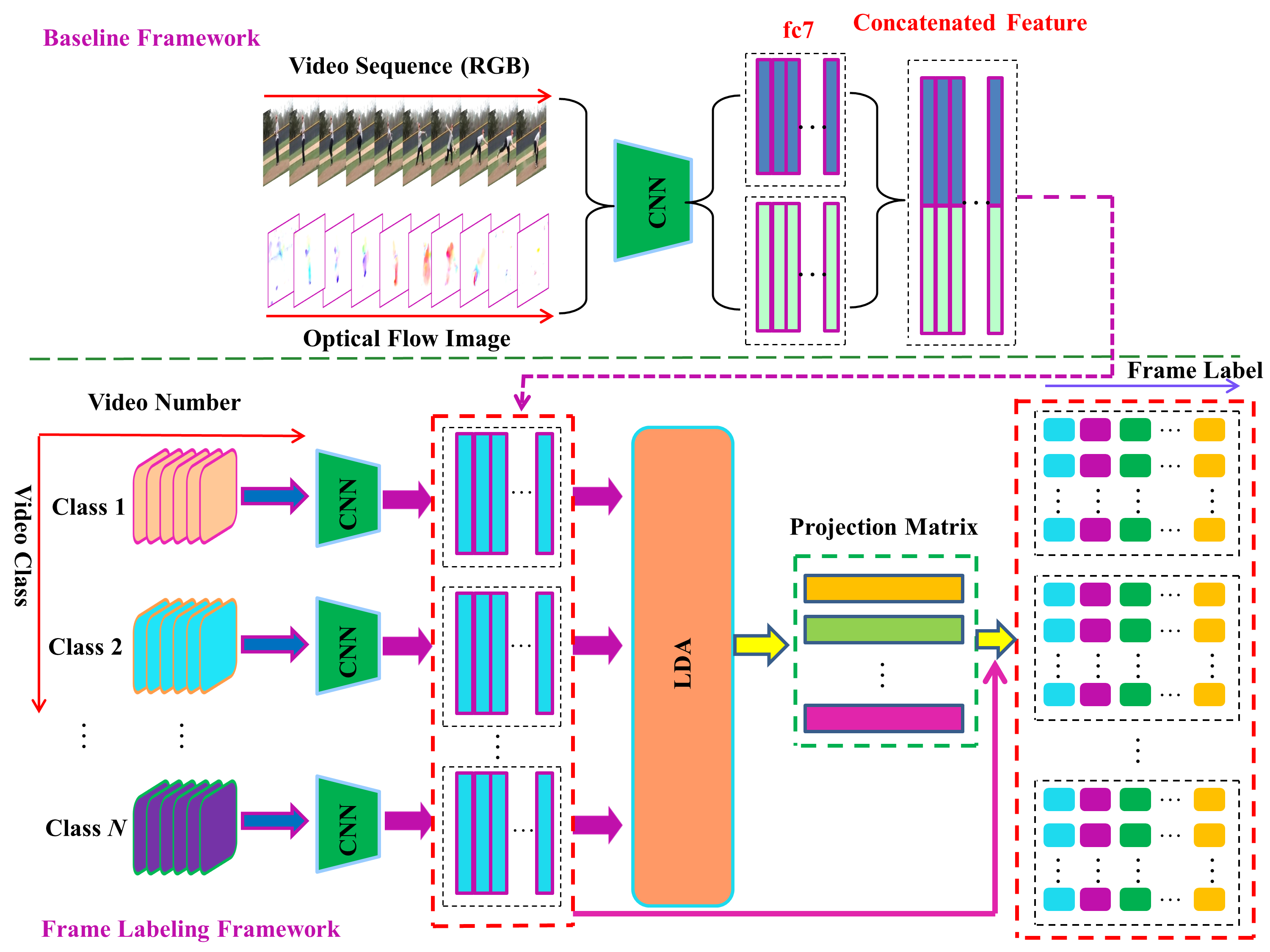}
\end{center}
\caption{ Automatically generating labels to train the deep keyframe detection framework. The appearance and motion information (the output of $ f_{c7} $ layer) are concatenated as a spatio-temporal feature vector. Thereon,  LDA is applied to all the feature vectors of all the classes of training videos to project  the feature vectors to a low dimensional feature space (LDA space) and obtain the projection vectors of each class videos. Finally, such projection vectors is applied to calculate the frame-level video labels.}
\label{fig:2}
\end{figure*}
%-------------------------------------------------------------------------
\section{Related Work}
\noindent {\bf Two-stream ConvNets} Two-stream ConvNets have achieved remarkable success in video analysis, especially for action recognition and action detection in videos~\cite{simonyan2014two,tran2015learning,wang2016temporal,feichtenhofer2016convolutional,feichtenhofer2016spatiotemporal, feichtenhofer2017spatiotemporal,Diba_2017_CVPR}. Karpathy \etal~\cite{karpathy2014large} trained a ConvNet with deep networking on a very large sports activities dataset (Sports-1M). However, the results obtained by this method were worse than the best hand-crafted representation method ~\cite{wang2013action} since it used the stacked video frames as input to the network and did not consider the temporal motion information of videos.
 To overcome this, Simonyan \etal~\cite{simonyan2014two} first proposed a two-stream ConvNet architecture which incorporated spatial and temporal
 networks and exploited an ImageNet challenge dataset for pre-training and calculating optical flow to explicitly capture motion information. More recently, Feichtenhofer~\etal~\cite{feichtenhofer2016convolutional,feichtenhofer2016spatiotemporal,feichtenhofer2017spatiotemporal} proposed three new two-stream ConvNets based on spatio-temporal multiplier architecture and residual networks. However, our model differs from these two-stream ConvNets by its appearance-motion feature aggregating for key frame detection in human action videos strategy, which formulates the key frame detection in human action videos as a regression problem, but not a common classification problem.

\noindent {\bf Key frame detection.} Many earlier approaches relied on using a segmentation based pipline. Such methods typically extracted optical flow and SIFT features ~\cite{wolf1996key,lowe2004distinctive,guan2013keypoint,kulhare2016key}. One of the first works~\cite{wolf1996key,kulhare2016key} described a video with optical flow and detected local minimum changes in terms of similarity between successive frames. Later works improved upon this pipline by using keypoints detection for feature extractions~\cite{lowe2004distinctive,guan2013keypoint}, which extracted the local features via a SIFT descriptor and pooled the keypoints to achieve key frames in videos. However, all of these approaches suffer the disadvantage that they may extract similar key frames when the same content reappears in a video.

Another class of methods relies on clustering the features (e.g. HS color histogram) of video frames into groups. These methods determine the key frames in a video by detecting a representative frame from each group. Zhuang~\etal~\cite{zhuang1998adaptive} proposed an unsupervised clustering method for identifying the key frame that takes into account the visual content and motion analysis. Cernekova~\etal~\cite{rasheed2005detection} improved upon the Cernekova~\etal~ by the mutual information (MI) and the joint entropy (JE) between consecutive video frames, by using the MI to measure information transported from one frame to another, and JE to exploit the inter-frame information flow respectively. Vazquez~\etal~\cite{vazquez2013spatio} proposed a spectral clustering based key frame detection method that built a graph to capture a feature's locality in an image video sequence instead of relying on a similarity measure computed by the features shared between two images. It is  worth mentioning here the works on sparse dictionary and  Multiple Instance Learning (MIL) by ~\cite{cong2012towards,mei2015video,meng2016keyframes,lai2014video,zhou2017real}. They used the dictionary learning or MIL framework to learn the features of video frames in order to detect the key frame for video summarization, action recognition or video event detection .

Due to the popularity of deep Convolutional Neural Networks (CNNs) in image classification, deep CNNs have been introduced in key frame detection in videos~\cite{yang2015unsupervised,Mahasseni_2017_CVPR,Kar_2017_CVPR,zhou2017real}. Yang~\etal~\cite{yang2015unsupervised} first introduced the bidirectional long short term memory (LSTM) for automatically extracting the highlights (key frames) from videos. Most recently, several deep learning based methods for key frame detection in videos have been proposed~\cite{Mahasseni_2017_CVPR,Kar_2017_CVPR}. Mahasseni~\etal~\cite{Mahasseni_2017_CVPR} first applied the Generative Adversarial Networks (GAN) to key frame detection in videos, which used CNNs to extract the feature of each frame and then encoded the feature via LSTM. Kar~\etal~\cite{Kar_2017_CVPR} adopted two-stream CNNS containing spatial and temporal net with the MIL framework to detect the key frame with high scores in videos for action recognition.

Our method differs from the architecture of these deep CNNs by automatically producing frame-level labels to train the deep key frame detection model, which can model using the entire video without the limitation of sequence length. The most important distinction is that we regard the compelling frames as key frames in human videos. In other words, our task is to report the detection results of compelling frames in videos. To the best of our knowledge, our work is the first to report the results of key frame detection via learning the two-stream ConvNets model. Moreover, our work is the first to automatically generate frame-level video labels via ConvNets.
%--------------------------------------------------------------
\section{Approach}
In this section, we give detailed description of performing key frame detection in videos with deep two-stream networks. Firstly, we illustrate our motivations and then, we present our deep key frame detection framework. Finally, we present our deep key frame detection framework. More details about our motivations and the key frame detection architecture are given in Section {\color{red} 3.1} and Section {\color{red} 3.2}.

\subsection{Motivation}
\begin{figure*}
\begin{center}
\includegraphics[width=0.95\linewidth]{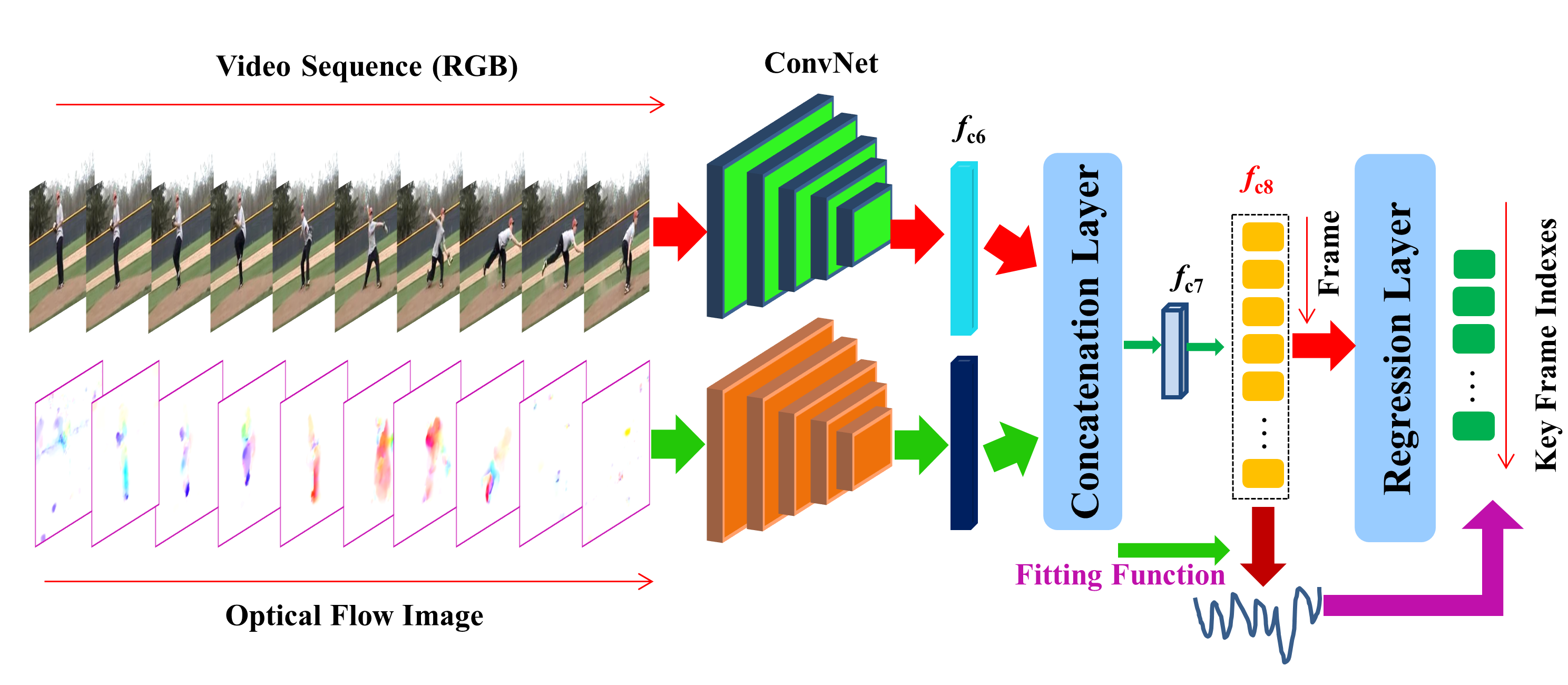}
\end{center}
   \caption{ An overview of the key frame detection two-stream ConvNets framework: appearance and motion networks. The appearance network operates on RGB frames, and the motion network operates on the optical flow represented as images. The appearance stream ConNets are  used to extract appearance information (the output of fc6-layer) and the motion stream ConvNets are used to extract the motion information (the output of fc6-layer). The feature maps from the appearance and motion ConvNets are aggregated to form a spatio-temporal feature representation fc7. Then the fc7 feeds into a fully connected layer(fc8) to form a feature vector with one value. Thereon,  the feature vector feeds into the regression layer to optimize predicted score. Finally, a smooth fitting function is applied to all the final predicted scores of the whole video to distinguish the key frames from the whole video sequence.}
\label{fig:3}
\end{figure*}

Generating effective video representations is relatively straight forward via the two-stream convolutional neural networks. Inspired by the key volume mining for action recognition and key frame detection for summarizing videos~\cite{zhu2016key,mahasseni2017unsupervised}, it is expected that our proposed key frame representation will reflect more semantic information extracted from the whole video. We propose a deep key frame detection approach, our goal is to discover the compelling frames in a human action video, that are able to discriminate against other classes containing different labeled category actions. To do this, we first label each frame of a video using the same label as the complete video since only video-level labels are generally available. Thereafter, we can use the frame-level labels and videos to learn a model that detects the key frame in a video. The proposed deep key frame detection architecture is illustrated in Figure~\ref{fig:3}.
%-------------------------------------------------------------------
%--------------------------------------------------
\subsection{Learning Key Frame Detection Model}

 To learn a key frame detection model, we need a good key frame discriminator; and to train a good key frame  discriminator, we need key frames labeled in videos. To do this, we first capture the frame-level video labels by fusing the video's      appearance and motion information. Then we make the frame-level video labels that must be applied to discriminately select the key frame in a video. The process requires that we first capture the optical flow images in two adjacent frames for each frame in the video, and then feed the RGB video sequence and its corresponding optical flow images to a CNN model, to extract the appearance and motion information of each video frame. Thereon, the appearance and motion features of each frame are fused to form a new feature representation, which can enforce the  representation capability of each frame. Subsequently, we apply the LDA to all the fused features of all the training videos to shrink them to a lower dimensional space to form a low dimensional feature representation. Such features will be used to produce the label (key frame information) of each frame. Once the label of each frame in a video is obtained, we will use the video sequences and their corresponding optical flow images to learn a two-stream ConvNets that can automatically generate the predictions of key frame locations in a video. The process of frame-level video labeling and that of key frame detection is as shown in Figures~\ref{fig:2}  and~\ref{fig:3}. Further details can be found below:

\vspace{2mm}
 \noindent {\bf Frame-level video labeling:} To label frame-level video, we propose a deep leaning based method as shown in Figure~\ref{fig:2}. We denote a video as $ X=\left [ \mathbf{x}_1, \mathbf{x}_1, \cdots,\mathbf{x}_K\right]$, $ \mathbf{x}_k\in \mathbb{R}^{224\times 224\times 3}$, with each frame $\mathbf {x}_k $ represented as RGB images. Assume a video \emph {X} belonging to labeled class \emph{c}, $c\in\left\{ 1,2,\cdots,C\right\}$, the output feature map from CNNs at a fully connected layer$(F_1,F_2,\cdots,F_K)$, and its output feature map from CNNs at the same fully connected layer of corresponding optical flow image is $(O_1,O_2,\cdots,O_K)$, where $ {F_k} $ and $ {O_k} $ are the CNN feature of \emph{k}-th frame RGB and its corresponding optical flow image. Similarly, we can obtain the RGB CNN features and the corresponding optical flow image CNN features of each  training video $ (V_{R,1},V_{R,2},\cdots,V_{R,N})$ and $(V_{O,1},V_{O,2},\cdots,V_{O,N})$, where \emph{N} denotes the number of training videos, $ V_{R,n} $ and $ V_{O,n} $ represent  RGB CNN features and its corresponding optical flow CNN features for \emph{n}-th training video. Specifically, $ V_{R,n} $ and $ V_{O,n} $ are composed of $(F_1,F_2,\cdots,F_K)$ and $(O_1,O_2,\cdots,O_K)$,  respectively.

 Thereon, RGB CNN features and optical flow CNN features are concatenated to fuse the appearance and motion information of each video to enforce the video representation ability. These  can be written as $ (V_{F1},V_{F2},\cdots,V_{FN}) $. To determine the difference between each frame in a video, we adopt the linear discriminant analysis (LDA)~\cite{prince2007probabilistic} technique to discriminate between videos from different classes because LDA can not only perform dimensionality reduction, but can also preserve as much of the class discriminatory information as possible. To address this issue, we aggregate the fused features of  all the videos belonging to the same class \emph {c} as a matrix $ V_c $, and all the features of video from the same class can be written as $ \left\{V_1,V_2,\cdots,V_C\right\}$. In this paper, we instruct LDA to classify a class \emph{c} from other classes. We regard the class \emph{c} to be distinguished as class \emph{A} and all the rest of the classes can be composed as a new class \emph{B}.
 For example, we compare class \emph{A} belongs to class 1, we can describe the feature matrix of class \emph{A} and class \emph{B} as follows:
 \begin{equation}
 V_A=V_1
\end{equation}
 \begin{equation}
 V_B  =\left[V_2,V_3,\cdots,V_C\right]
 \end{equation}
 Given $ V_A $ and $ V_B $, we can perform LDA to obtain the projection vector $ W_A $,
 \begin{equation}
 W_A={\rm LDA}(V_A,V_B)
 \end{equation}
 Then we  use $ W_A $ to calculate each frame score (label value) for each training video of class \emph{A}:
 \begin{equation}
 f_{i,m}=\left\|F_{i,m}-W_A^{T}F_{i,m}\right\|_{2}
 \end{equation}
 Where $ F_{i,m} $ represents the feature vector of the \emph{i}-th frame of the \emph{m}-th video of class \emph{A}.  Similarly, we can obtain the feature score of each frame in each video of each class for all the training videos via Equation (4).

\vspace{2mm}
\noindent {\bf Key frame detection model learning :} Given the label $ f_{i,m} $ of each frame in each training video obtained above, the training videos and corresponding optical flow images are applied to learn our key frames selection model. Our proposed model is a deep two-stream convolutional neural networks architecture. Specifically, one stream is appearance stream A and the other is motion stream B. The architecture of these two streams is similar to the VGG-16 network, except for the removal of the last layers (fc7, fc8, and softmax) from the second fully connected layer (fc6). Based on fc6-a ( fc6 for CNN stream A) and fc6-b (fc6 for CNN stream B), the input of the new fully connected layer fc7  is formed by concatenating fc6-a and fc6-b, then feeding the output of the fc7 to fc8. The final layer is a regression layer instead of a softmax layer.
   %--------------------------

\begin{figure}
\begin{center}
\includegraphics[width=1.0\linewidth]{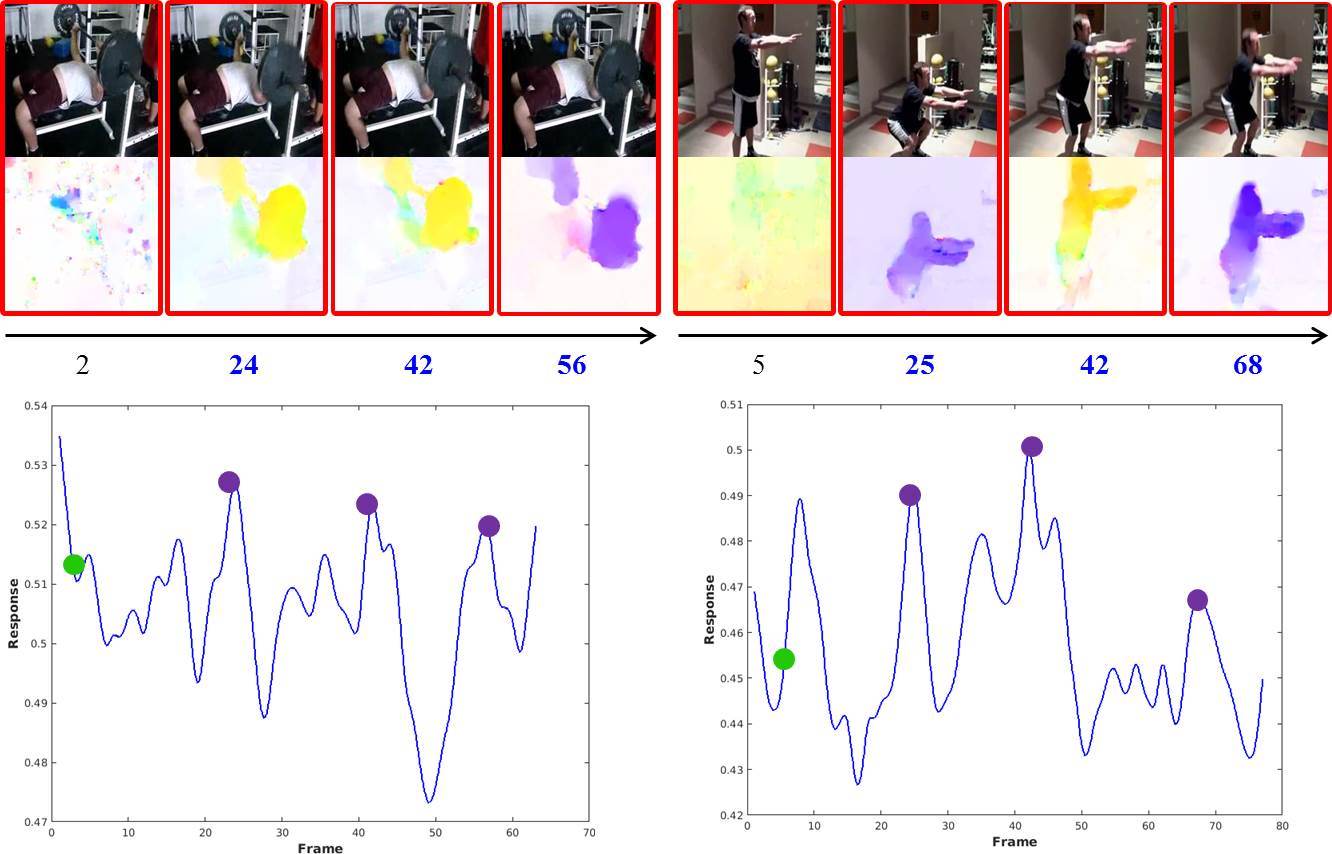}
\end{center}
\caption{Example responses of our network and the corresponding ground truth curves obtained by LDA learning for two UCF101 videos. The first two rows show RGB sequence frames and their corresponding optical flow images. The last row shows the ground truth curves. The purple solid circle represents the key frames.}
\label{fig:6}
\end{figure}

    Specifically, assume the output of the RGB and optical flow feature maps of the CNNs of each video is $\left\{F_1,F_2,\cdots,F_M\right\}$ and $\left\{F^{\left(1\right)},F^{\left(2\right)},\cdots,F^{\left( M \right)}\right\}$. Then the appearance-motion aggregated features are given by $ \left\{G_1,G_2,\cdots,G_M\right\} $. And the $ G_m $ is obtained by concatenating the RGB and optical flow features. After concatenation, we obtain a new fully connected layer feature, and we adopt the $ L_2 $ regression loss function to minimize the differences between the estimated and the ground truth target values. Here, the prediction outputs of a video \emph{V} are written as $ \left\{S_1,S_2,\cdots,S_M\right\} $. Thereon, we use these prediction outputs of a video to determine the key frames of this video. To do this, we use the \emph{smoothing spline} function to fit a curve to the prediction outputs of a video. In Equation (4), if the frames are very similar to the frames belonging to the same category and different from the frames of other categories, they belongs to the key part of action. However, for the parts of action that occur with sudden transition, they have higher response compared to the surrounding frames. Hence the local maximum between the two minimum corresponds to fast moving activities in the fitted curve on the video sequence frame and the minimum responses correspond to small changes in the activities. Frames are dynamically selected depending on the content of the video. Hence, complex activities or events would have more key frames compared to simpler activities. Example responses of our network and the corresponding ground truth curves obtained by LDA learning for two UCF101 videos is shown in Figure~\ref{fig:6}.
%------------------------------------------
%------------------------------------------
\subsection {Implementation Details}
We use the Caffe toolbox~\cite{jia2014caffe} for ConvNet implementations and all the networks are trained on one  NVIDIA  GPU. Here, we describe the implementation details of our frame-level video annotation scheme and key frame detection scheme.

 \noindent {\bf Frame-level video labeling:} In our experiments, we use the VGG-16 network~\cite {simonyan2014very} trained on the ImageNet dataset~\cite{deng2009imagenet} to extract the appearance features and motion features from the input video frames and their corresponding optical flow images.  fc7-layer features of video frames and the optical flow images are extracted and concatenated into a 8192-dimensional visual feature vector. Such a vector of each frame in each video is used to label each individual frame.
 
 \noindent {\bf Two-stream ConvNets:} We employ the pre-trained VGG-16 ~\cite {simonyan2014very} model trained on the ImageNet dataset~\cite{deng2009imagenet} to design the two-steam ConvNets. The two-stream network consists of appearance and motion networks. The appearance ConvNet operates on RGB frames and, the motion ConvNet operates on dense optical frames which are produced from two adjacent RGB frames. The input RGB image or optical flow frames are of size $ 256\times 340 $, and are central cropped to a size $ 224\times 224 $, and then mean-subtracted for network training. To fine-tune the network, we replace the fc6 layer with a concatenated layer, (which is obtained by aggregating the fc6 of appearance ConvNet and motion ConvNet) and replace the fc7 with an X-length layer, and the previous classification layer with a one-way regression layer. The softmax layer is replaced by a Euclidean loss layer. Moreover, we use the video sequence RGB frames and their corresponding optical flow images as training data to fine-tune the original network instead of the ImageNet dataset. We use mini-batch stochastic gradient descent (SGD) to learn the network parameters, where the batch size is set to 16 and momentum set to 0.9.
 
 In our experiments while training the model, we retain only the  convolutional layers and the first fully connected layer (fc6) of each convolutional layers in each network. More specifically we remove the other fully connected layer, classification layer and softmax layer. The feature maps extracted from the first fully connected layer from each network are fused as a new feature map, which feeds into the next  fc7  layer in our model. For example, both of the two feature maps are the outputs of the first fully connected layer of VGG-16 with the size  $ 4096\times 1 $, leading to a new fc6  layer feature with the size $ 8192\times 1 $. We fine-tune the whole model to train spatial and temporal joint ConvNets, we then initialize the learning rate with $ 10^{-3} $ and decrease it by a factor of 10 for every 1,600 iterations. We rescale the optical flow fields linearly to
a range of [0, 255]. For the extraction of the optical flow images, we use the TVL1 optical flow algorithm ~\cite{zach2007duality} from the OpenCV toolbox. We use batch normalization.

\noindent {\bf Fully-connected pooling and regression layer:} In order to efficiently represent video and  encode the appearance and motion features,
 we proposed to first concatenate the first fully-connected layer (fc6-a and fc6-b) feature vectors of the two streams to form a 8192-dimensional ($ 4096\times 2 $) feature vector and then pass them through a fully-connected layer consisting of 1024 units, thereon, this 1024-dimensional feature vector passes through another fully-connected layer consisting of only a 1-dimensional unit. Thus, the proposed model extracts a 1-dimensional feature vector for every frame in an action video. Based on this, we fine-tuned the last two fully-connected layers (fc7 and fc8) between the fused layer and the regression layer for the fusion ConvNets, we initialized the learning rate with $ 10^{-2} $ and decrease it by a factor of 10 for every 3400 iterations in the model training step. We ran the training for 10 epochs for both the two stream networks. For the regression layer training, we
choose the Euclidean  layer for loss function.

\noindent{\bf Fitting function setting:} In our experiments, we use the \emph {smoothing spline} function for fitting a curve using the predicted values of each frame to find out the key frames. This function, has a parameter $ \alpha $ that controls the fitting shape. We set $ \alpha=0.8 $ when the frame number in a video is less than 60 and $ \alpha=0.1 $ if the frame number in a video is more than 170. Otherwise, we set $\alpha=0.6 $. Empirically, we find that this setting demonstrates good performance. However, other settings may obtain equally good results.

\section{Experiments}
In this section, we first conduct experiments to validate the proposed framework and novel techniques, and then analyse  the results for different category videos.
 %-------------------------------------------------------------------------
\begin{table}[h]
 \centering
 \begin{tabular}{| l | c | c |}
 \hline
  \multicolumn{3}{|c|}{\bf Average Error in Keyframe}\\
 \hline
  Class Name & Number & Position \\ \hline
  \hline
  Baseball Pitch & $\pm 1.73 $ & $\pm 0.455 $  \\ \hline
  Basket. Dunk   & $ \pm 1.64 $ & $ \pm 0.640 $\\ \hline
  Billiards      & $ \pm 2.18 $ & $ \pm 1.300 $ \\ \hline
  Clean and Jerk & $ \pm 2.27 $ & $ \pm 1.202 $ \\ \hline
  Cliff Diving   & $ \pm 2.45 $ & $ \pm 0.627 $ \\ \hline
  Cricket Bowl.  & $ \pm 2.45 $ & $ \pm 1.14 $ \\ \hline
  Cricket Shot   & $ \pm 1.27 $ & $ \pm 0.828 $ \\ \hline
  Diving         & $ \pm 2.00 $ & $ \pm 0.907 $  \\ \hline
  Frisbee Catch  & $ \pm 1.73 $ & $ \pm 0.546 $ \\ \hline
  Golf Swing     & $ \pm 2.45 $ & $ \pm 0.752 $ \\ \hline
  Hamm. Throw    & $ \pm 1.73 $ & $ \pm 1.223 $\\ \hline
  High Jump      & $ \pm 1.73 $ & $ \pm 0.434 $ \\ \hline
  Javelin Throw  & $ \pm 2.45 $ & $ \pm 0.555 $ \\ \hline
  Long Jump      & $ \pm 2.27 $ & $ \pm 0.611 $\\ \hline
  Pole Vault     & $ \pm 2.64 $ & $ \pm 1.139 $\\ \hline
  Shotput        & $ \pm 2.00 $ & $ \pm 0.564 $\\ \hline
  Soccer Penalty & $ \pm 2.09 $ & $ \pm 0.712 $\\ \hline
  Tennis Swing  & $ \pm 1.64 $ & $ \pm 0.554 $ \\ \hline
  Throw Discus   & $ \pm 2.09 $ & $ \pm 0.642 $ \\ \hline
  Volley. Spike  & $ \pm 1.54 $ & $ \pm 0.633 $\\ \hline
  \hline
  {\bf Average accuracy} & $ \pm \bf {2.02} $ & $ \pm \bf{0.773} $\\ \hline
  %\multicolumn{3}{| l |}{{\bf Average accuracy}} & z \\ \hline
  \end{tabular}
  \caption{Accuracy performance comparison of the location and number matching of key frames obtained by our learning model with the ground truths using 20 classes of the UCF101 dataset.}
  \label{table:tabel1}
  \end{table}
%-----------------------------------------------------------------------
\begin{figure*}
\begin{center}
\includegraphics[width=0.95\linewidth]{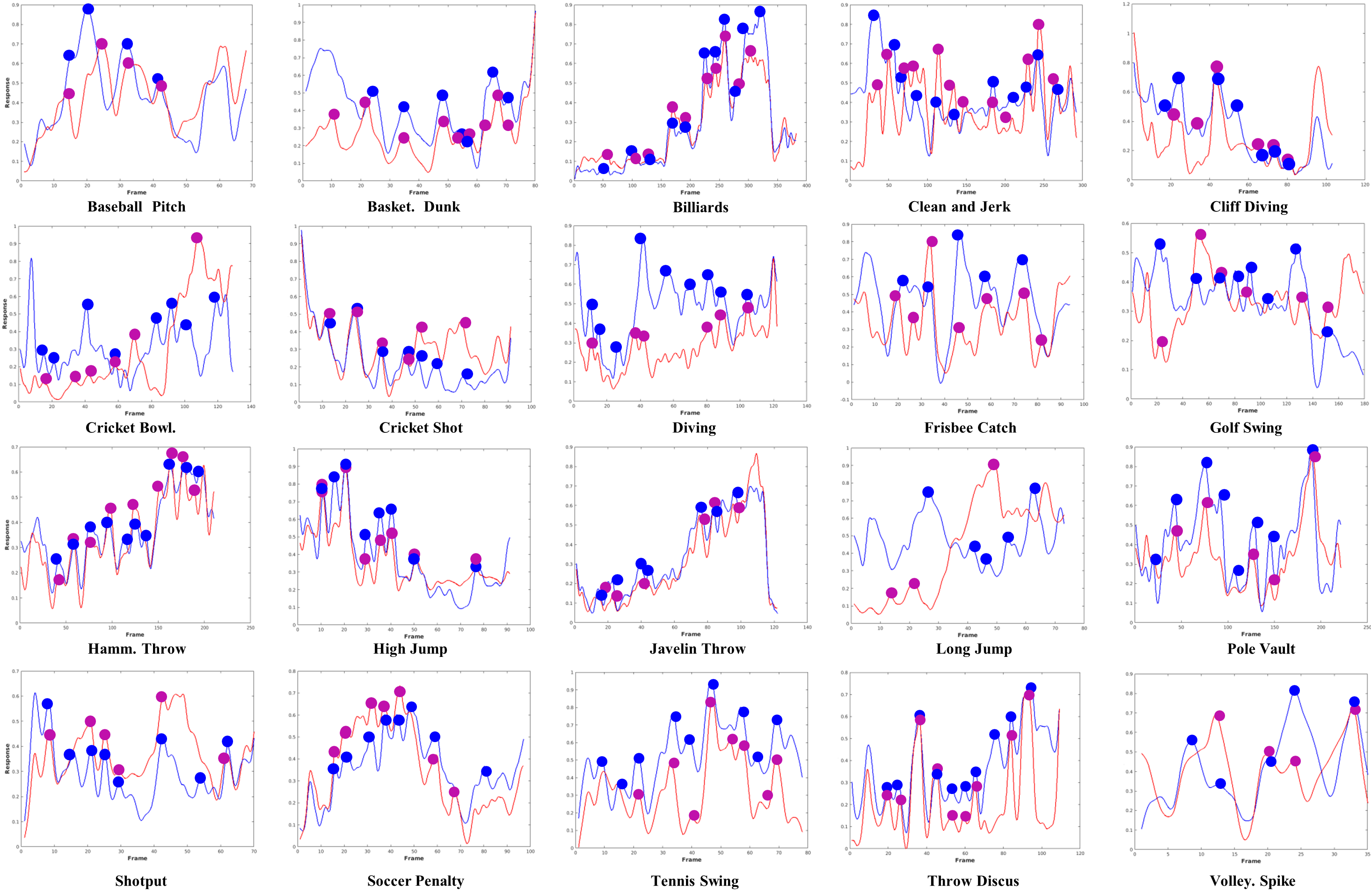}
\end{center}
\caption{ Example of the key frame detection results and ground truths. The blue curve represents the ground truth curve and the red curve represents the detected result curve. The blue solid circle represents the temporal location of ground truth and the  purple solid circle represents the key frame temporal location of the detected result.}
\label{fig:4}
\end{figure*}
%-------------------------------------------------------------------------
\subsection{Dataset}
Since there is no benchmark dataset for key frame detection in human action videos, we conduct our key frame detection experiments on a challenging dataset for video based human action recognition, namely UCF101~\cite{soomro2012ucf101}. This dataset consists of 101 actions classed with 13,320 video clips. It has at least 100 video clips for each action category. Besides its 101 categories, UCF101 has coarse definitions which divide the videos into human and object interaction, human and human interaction and sports. Although, this dataset is designed for human action recognition, we use it to perform key frame detection experiments. However, for key frame detection, we just use 20 classes videos (200 videos)  to train the data to learn the key frame detection model. To test the performance of our model, we randomly select 300 videos with the same categories for training, where each class has 15 videos.
 %--------------------------
\begin{figure*}
\begin{center}

\includegraphics[width=0.95\linewidth]{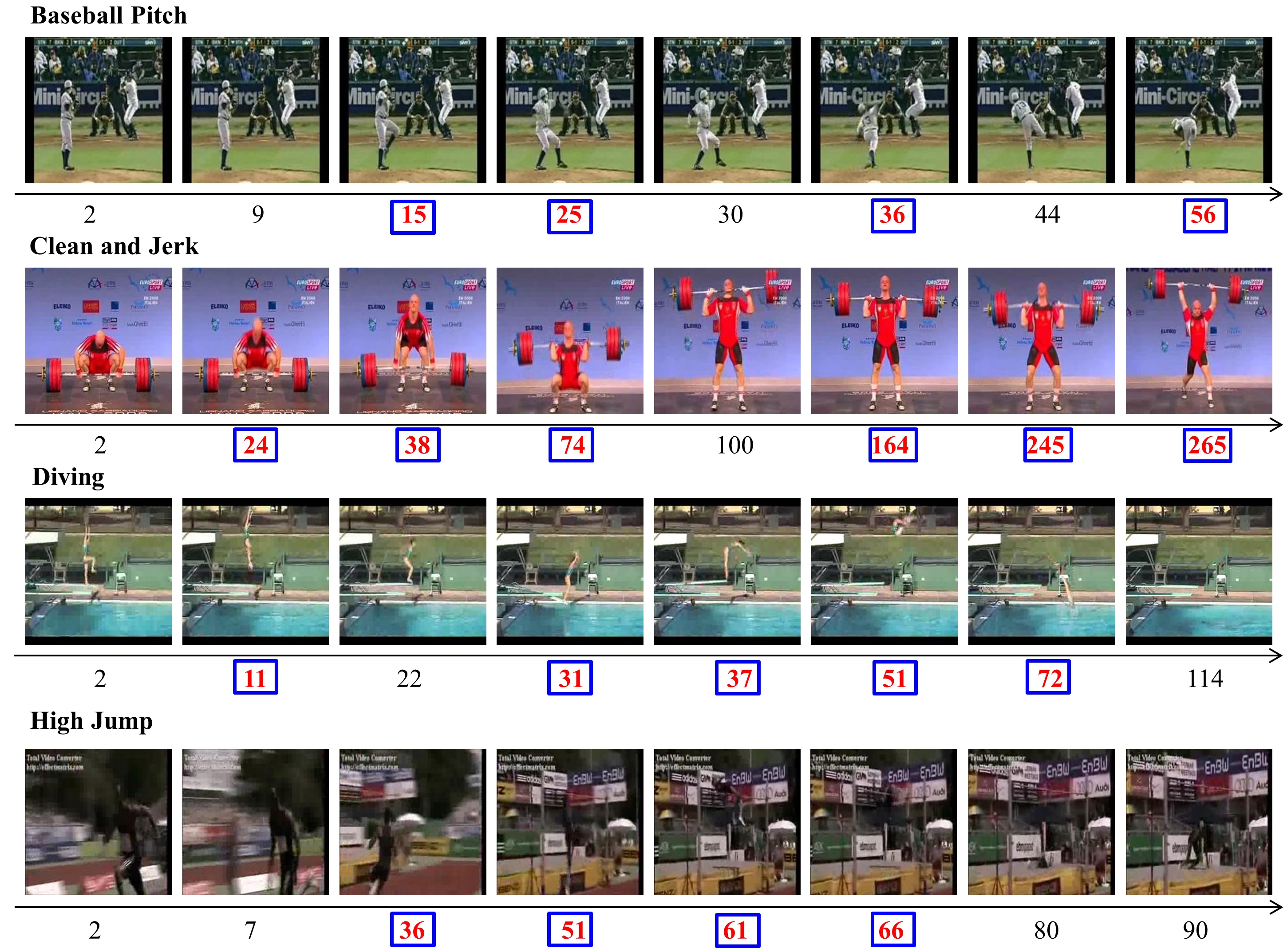}
\end{center}
\caption{ Visualizations of key frames detection. The  red frames surrounded with blue rectangular box represent the detected key frames. The timeline gives the position of the frame as a percentile of the total number of frames in the video (best seen in colour).}
\label{fig:5}
\end{figure*}
%--------------------------------------------------------------------
\subsection{Metrics}
As there are no benchmarks or ground truth results for key frame detection algorithms so far, it is not possible to present comparisons with any other method. Before we can evaluate our experimental results, we need to define some meaningful evaluation metrics as our objective is to match the ground truth. One direct measure would be evaluate how many frames predicted key frames that differ from the ground truth in the test data. Another measure would be to evaluate the temporal distance between the location of the key frame and the ground truth. Since we have both the number and location information of the key frames, in the following, we derive the two types of evaluation metrics accordingly.

\noindent{\bf Key frame number match error:} Detecting too many irrelevant frames as key frames in a video would undermine the main objective of video summarization. It will also affect subsequent processing in terms of more computations and less accuracy due to insignificant key frames. Thus, we propose a key frame number match error $ E_n $ to evaluate the detection accuracy. We first assume the ground truth number of key frame for a given video is $ G_n $ and the predicted number of key frame is $ P_n $, then, the number matching error can be described as:
 \begin{equation}
 E_n= P_n-G_n
\end{equation}

\noindent{\bf Key frame location match error:} The temporal location matching of key frames is very important in key frame detection, since it directly illustrates whether the detected frame is indeed a key frame i.e. important for video summarization. We assume the ground truths of key frame locations are $ \left\{G_{l}^{\left(1\right)}, G_{l}^{\left(2\right)},\cdots, G_{l}^{\left(k\right)}\right\} $ and the predicted locations are $ \left\{P_{l}^{\left(1\right)}, P_{l}^{\left(2\right)},\cdots, P_{l}^{\left(k\right)}\right\} $. Then the location match error is described as:
\begin{equation}
 E_l=\pm \frac{1}{k}\sum_{x=1}^{k}\left|P_{l}^{\left(x\right)}-G_{l}^{\left(x\right)} \right|
\end{equation}
%---------------------------------------------
\subsection{ Qualitative Results}
In this part, we apply our trained two-stream ConvNets to detect key frames in videos. We report the performance of our proposed method for key frame detection and give qualitative results for different category videos. Since there is no literature that reports the results of key frames detection for the UCF101 dataset, we just compare our results with the ground truth and validate our method's performance. Table~\ref{table:tabel1} compares the number and location errors between the detected key frames and ground truth. It can be clearly seen that our proposed method obtained low error for number and location of key frames, and the average location error is lower than one frame. Figure~\ref{fig:4} shows a visualization of the results of key frame detection and the ground truths for videos of different category. From Figure~\ref{fig:4}, we can also see the difference between the detected results and ground truths for both numbers and locations. Moreover, Figure~\ref{fig:5} shows some typical cases (four test videos) visualized with the output from the proposed deep keyframes detection algorithm. In the ``Baseball Pitch'' example we observe that our deep model detects the key frames that seem to correspond to (i) the movement of lifting the leg, (ii) throwing the baseball and (iii) the bend of the body after the throw. We also see a similar trend in the ``Clean and Jerk'' example, where deep key frame detection model detects the frames corresponding to (i) the initial preparation, (ii) the lifting of the barbell and (iii) dropping the barbell. In the ``Diving'' example, the detected key frames correspond to (i) taking off, (ii) flipping in the air, and (iii) diving into the water. In the ``High Jump'' example, the detected key frames correspond to (i) running, (ii) taking off, and (iii)jumping over the pole and landing.  Such frame detections resonate with previous works that have highlighted the general presence of three atomic actions in classes that can describe certain actions~\cite{gaidon2013temporal}. These visualizations strengthen our claim that our deep ConvNet is able to well and truly distinguish between human actions captured in video frames by detecting the differences in each frame. We further observe from these visualizations that our deep ConvNets also implicitly learn to decompose actions from certain classes into simpler sub-events. In other words, our deep ConvNets can output key frames which can correctly represent a video shot with less redundancy in frames.

\section{Conclusion}
We presented a key frame detection deep ConvNets framework which can detect the key frames in human action videos. Such frames can correctly represent a human action shot with less redundancy, which is helpful for video analysis task such as video summarization and subsequent action recognition.  The proposed method used to address this issue was learning to dynamically detect key frames for different videos. However, there is no benchmark dataset for evaluating key frame detection methods. To overcome this, we used the popular action recognition dataset UCF101 to perform our experiments. However, this dataset lacks frame-level labels. To tackle this issue, we employed an automatically generating frame-level label approach by combining the CNN feature of each frame and LDA. Then a deep ConvNets for key frames detection model was learned by combining the RGB video frames and their corresponding optical flow images.  We verified our method on the dataset UCF101 and obtained encouraging results. This work is the first to report results for key frame detection in human action videos via deep learning.
%-------------------------------------------------------------------------
{\small
\bibliographystyle{ieee}
\bibliography{egbib}
}
\end{document}